\documentclass[10pt,twocolumn,letterpaper]{article}

\usepackage{cvpr}              

\usepackage{adjustbox}
\usepackage{tabularx}
\usepackage{multirow}
\usepackage{tikz}
\usepackage{algorithm}
\usepackage{algpseudocode}
\usepackage{enumitem}
\usepackage{caption}
\usepackage{graphicx}
\usepackage{wrapfig}
\usepackage{amsmath}
\usepackage{url}
\usepackage{array}
\usepackage{soul, color, xcolor}
\usepackage{setspace}

\definecolor{cvprblue}{rgb}{0.21,0.49,0.74}
\usepackage[pagebackref,breaklinks,colorlinks,allcolors=cvprblue]{hyperref}

\def\paperID{16842} 
\def\confName{CVPR}
\def\confYear{2025}

\title{Think Small, Act Big: Primitive Prompt Learning \\
for Lifelong Robot Manipulation}

 \author{
        Yuanqi Yao$^{1,7}$\thanks{Equal contribution: \href{mailto:yaoyuanqi@pjlab.org.cn}{yaoyuanqi@pjlab.org.cn}.}  \hspace{1.5em} 
        Siao Liu$^{2}$\footnotemark[1] \hspace{1.5em} 
        Haoming Song$^{1,3}$\footnotemark[1] \hspace{1.5em}
        Delin Qu$^{1,2}$ \hspace{1.5em}
        Qizhi Chen$^{1,4}$ \\
        Yan Ding$^{1}$ \hspace{1.5em}
        Bin Zhao$^{1,5}$ \hspace{1.5em}
        Zhigang Wang$^{1}$ \hspace{1.5em}
        Xuelong Li$^{6}$  \hspace{1.5em}
        Dong Wang$^{1}$\thanks{Corresponding author: \href{mailto:dongwang.dw93@gmail.com}{dongwang.dw93@gmail.com}.} \\
        $^{1}$Shanghai AI Laboratory \hspace{0.6em}
        $^{2}$Fudan University \hspace{0.6em}
        $^{3}$Shanghai Jiao Tong University \hspace{0.6em}
        $^{4}$Zhejiang University \hspace{0.6em} \\
        $^{5}$Northwestern Polytechnical University \hspace{0.6em}
        $^{6}$TeleAI, China Telecom Corp Ltd \hspace{0.6em}
        $^{7}$INSAIT, Sofia University 
}

\begin{document}
\maketitle
\begin{abstract}
Building a lifelong robot that can effectively leverage prior knowledge for continuous skill acquisition remains significantly challenging. Despite the success of experience replay and parameter-efficient methods in alleviating catastrophic forgetting problem, naively applying these methods causes a failure to leverage the shared primitives between skills. To tackle these issues, we propose Primitive Prompt Learning (PPL), to achieve lifelong robot manipulation via reusable and extensible primitives. Within our two stage learning scheme, we first learn a set of primitive prompts to represent shared primitives through multi-skills pre-training stage, where motion-aware prompts are learned to capture semantic and motion shared primitives across different skills. Secondly, when acquiring new skills in lifelong span, new prompts are concatenated and optimized with frozen pretrained prompts, boosting the learning via knowledge transfer from old skills to new ones. For evaluation, we construct a large-scale skill dataset and conduct extensive experiments in both simulation and real-world tasks, demonstrating PPL's superior performance over state-of-the-art methods. 

\end{abstract}

\vspace{-10pt}

\section{Introduction}

\begin{figure}[ht]
    \vspace{10pt}
    \centering
    \includegraphics[width=0.9\linewidth]{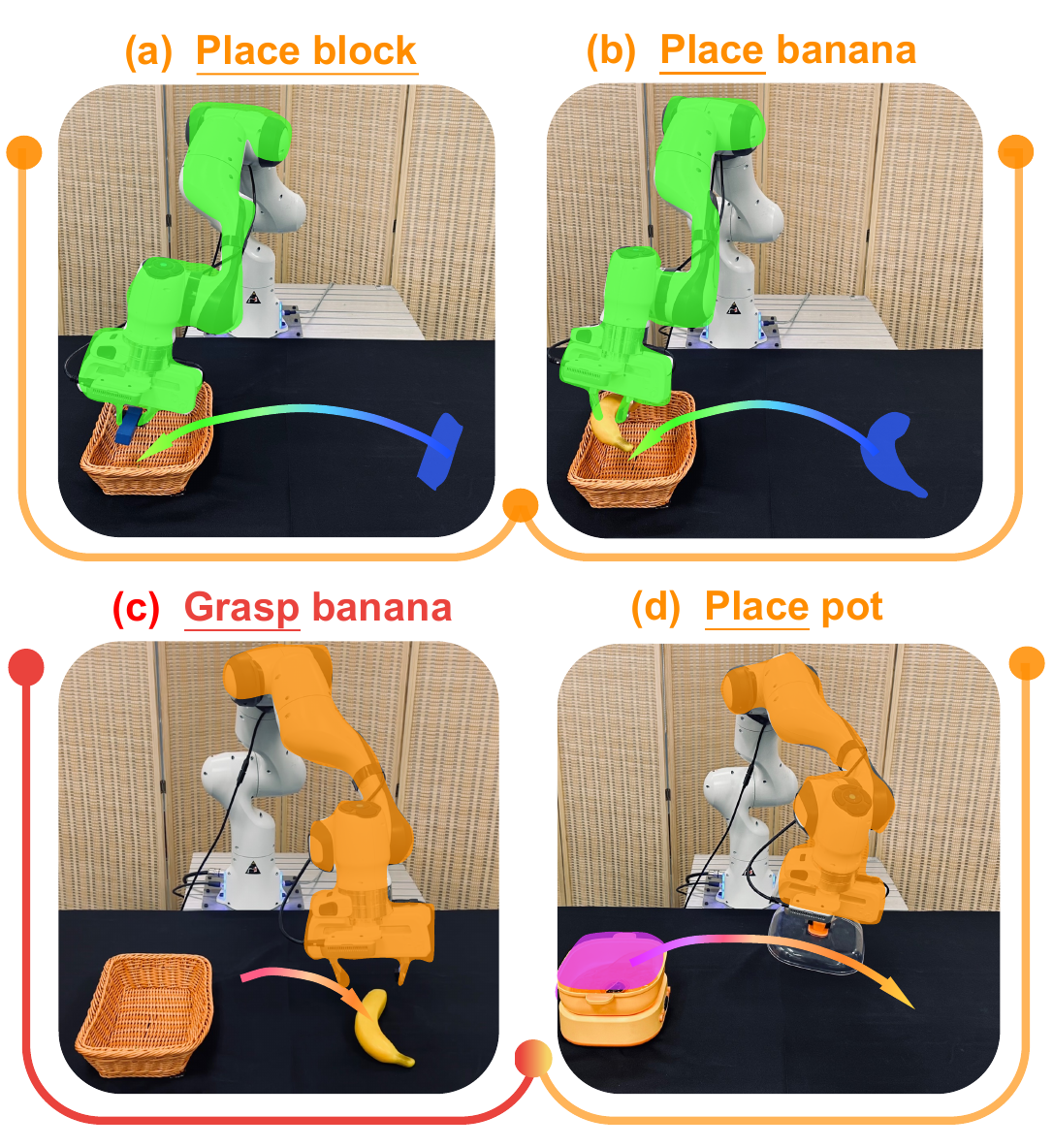}
    \vspace{-5pt}
    \caption{Optical flow captures primitive-level motion patterns, revealing latent shared knowledge between semantically similar skills $(a, b)$ and distinct skills $(c, d)$.}
    \label{fig:motivation}
    \vspace{-20pt}
\end{figure}

Developing robots that can effectively leverage prior knowledge to continuously acquire new skills has been a longstanding goal of generalist robotics.
Humans demonstrate remarkable abilities in utilizing past experiences to accelerate new learning - for instance, a person who knows how to use chopsticks can quickly adapt this knowledge to learn using tweezers, transferring their understanding of precision gripping and object manipulation across similar tasks.
However, implementing effective knowledge transfer in robotics lifelong span proves particularly challenging due to the complexity and diversity of robotic tasks.


\begin{figure*}[t]
\centering
\includegraphics[width=1.0\textwidth, keepaspectratio, interpolate=true]{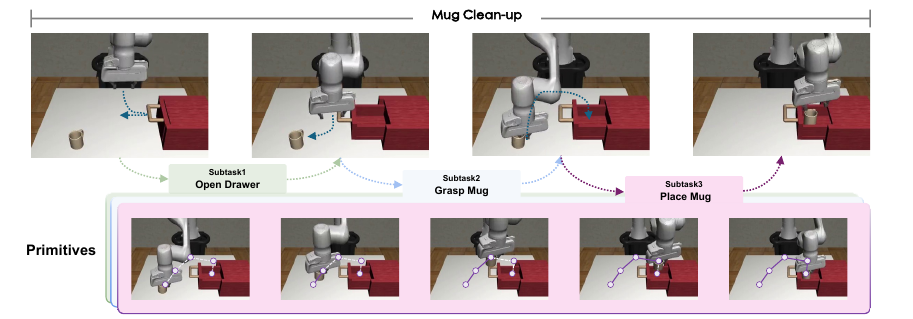}
 \vspace{-15pt}
\caption{\textbf{Illustration of primitives.} We demonstrate the concept of primitives within a task from MimicGen, focusing on "Mug Clean-up," which is composed of three subtasks: "open-drawer," "grasp mug," and "place mug." The bottom row provides a schematic representation of the primitives' trajectory, conceptualizing the subsequences of states. By leveraging the primitive prompts from our Primitive Prompt Learning (PPL) framework, we model the shared primitives from demonstrations, ultimately facilitating lifelong knowledge transfer. }
\vspace{-10pt}
\label{fig:primitives}
\end{figure*}

To facilitate effective knowledge transfer for lifelong skill acquisition, various approaches have been proposed. 
Some works explore experience replay ~\cite{rolnick2019experience, sodhani2020toward} to explicitly utilize prior collected data, but these methods face practical limitations in real-world deployments due to memory constraints and privacy concerns.
Other approaches leverage regularization or dynamic architectures~\cite{kirkpatrick2017overcoming,zenke2017continual,li2017learning}, aiming to better maintain existing knowledge when learning new tasks. However, these methods often yield suboptimal performance when scaling to complex vision-based manipulation tasks~\cite{aljundi2018memory,serra2018overcoming}, as they lack useful knowledge transfer mechanisms.
More recently, TAIL~\cite{liu2023tail} explored using Low-Rank Adaptation (LoRA~\cite{hu2021lora}) 
for each task, which prevents interference between different tasks.


Despite these advancements, a common limitation across existing works is their failure to effectively leverage shared knowledge between skills, limiting their ability to enable truly lifelong learning in robotics. As shown in Fig.~\ref{fig:motivation}, skills like "Grasp banana" and "Place pot", while semantically different, may share common underlying motion primitives. Fig.~\ref{fig:primitives} further illustrates this concept, demonstrating how a complex task like "Mug Clean-up" can be decomposed into subtasks, which in turn consist of more fundamental primitives. These primitives, representing basic motion patterns, form the building blocks that enable knowledge transfer across diverse robotic skills.

Recognizing and leveraging these shared primitives is crucial for effective knowledge transfer and lifelong learning across diverse robotic skills. Some prior works have explored various approaches for learning primitives from demonstrations, including pre-defined primitive functions and statistical models like hidden Markov models~\cite{pastor2009learning,manschitz2015learning,kroemer2015towards,chitnis2020efficient,fox2019hierarchical}, but such methods can be restrictive in representing complex dynamics.
More recent works have employed pre-trained deep neural networks to identify primitives from input states~\cite{gams2018deep,seker2019conditional}, while others have developed parameterized behavior primitives for specific manipulation tasks, such as movement primitives~\cite{ijspeert2013dynamical,neumann2014learning}, motion planning~\cite{garrett2021integrated,lozano2014constraint,toussaint2015logic,jia2024chain}, and grasping systems~\cite{bohg2013data,mahler2017dex}. Hierarchical imitation learning~\cite{yu2018one,le2018hierarchical,kipf2019compile,xie2020deep} has also shown promise in learning primitives from demonstrations.
These works have already revealed the benefits of primitives, such as reusability, modularity, and robustness to variations~\cite{gao2024prime}, However, they have not explored how to better leverage primitives to address lifelong robot learning challenges.


In this paper, we propose \textbf{P}rimitive \textbf{P}rompt \textbf{L}earning (\textbf{PPL}) for lifelong robot manipulation, a novel two-stage framework that transfers knowledge across skills through reusable and extensible primitive prompts. 
In the multi-skill pre-training stage, our framework learns a set of primitive prompts to represent shared and reuseable primitives. To achieve this, we design a motion-aware prompting via multi-modal text-flow queries, aiming to simultaneously capturing semantic and motion shared primitives across skills. Specifically, the motion-aware prompt is represented by weighted-sum of shared prompts and prepended into the keys and values of multi-head self-attention layers of diffusion transformer-based policy. 
In the lifelong learning stage, new lifelong prompts are appended and optimized with frozen pretrained prompts, enabling knowledge transfer between old and new skills. 
To evaluate PPL, we construct a large-scale skill dataset and conduct extensive experiments in both simulation and real-world tasks, demonstrating significant performance improvements over state-of-the-art methods. Our contributions are as follows:
\begin{itemize}[noitemsep, topsep=0pt, partopsep=0pt, parsep=1pt, itemsep=1pt, leftmargin=*]
    \item[$\bullet$] We propose Primitive Prompt Learning (PPL), a novel two-stage framework tailored for lifelong robot manipulation via reusable and extensible primitive prompts.
    \item[$\bullet$] We present motion-aware prompting via text-flow query, designed to represent shared primitives between skills and effectively transfer them to new skill acquisition.
    \item[$\bullet$] We construct a large-scale skill dataset and extensive experiments in both simulated and real-world environments, demonstrating significant performance improvements over state-of-the-art methods in lifelong robotic manipulation.
\end{itemize}


\section{Related Work}
\noindent \textbf{Lifelong Learning.}   
Lifelong learning for decision-making strives to create agents capable of continuous learning from ongoing data streams without suffering catastrophic forgetting~\cite{parisi2019continual,lesort2020continual,khetarpal2020towards,wan2024lotus}. Techniques such as experience replay~\cite{rolnick2019experience,shin2017continual,van2020brain} help maintain knowledge but impractical with increasing tasks due to growing memory demands. To counter this, methods involving regularization and dynamic architectures have been explored~\cite{kirkpatrick2017overcoming,zenke2017continual,li2017learning,liang2022search,cheng2023league}, yet they often yield suboptimal performance due to their indirect approach to knowledge retention~\cite{aljundi2018memory,serra2018overcoming}. More recently, TAIL~\cite{liu2023tail}, integrated with LoRA~\cite{hu2021lora}, demonstrates state-of-the-art results in lifelong learning with minimal parameter updates, but it requires specific parameters for each task and does not leverage learned knowledge to boost new skill acquisition, making it inefficient in the real world.


\noindent \textbf{Parameter-Efficient Methods for Lifelong Learning.}
Parameter-efficient methods have achieved significant progress in natural language processing~\cite{qin2021lfpt5,zhang-etal-2022-continual,qin-etal-2023-lifelong,zhao-etal-2024-sapt,wang-etal-2023-orthogonal} and computer vision~\cite{Hu2022,Wang2022,Jiang2023,yan2023gs,qu2023implicit} as a crucial way for adapting to new tasks in lifelong learning without substantially increasing parameters. 
Dual-Prompt~\cite{Wang2022} combines fixed and learnable prompts for knowledge preservation and task adaptation.
Coda-Prompt~\cite{Jiang2023} decomposes prompts into task-shared and task-specific components for efficient knowledge transfer. $\text{MP}^2$~\cite{Hu2022} leverages the task IDs to select prompts for few-shot learning. However, these methods face limitations in robot manipulation: Dual-Prompt and Coda-Prompt struggle with temporal dependencies, while $\text{MP}^2$'s task ID-based approach limits generalization across different robotic motions and tasks.

\noindent \textbf{Learning with Robotic Primitives.}
Prior work on robotic primitives learning can be broadly categorized into two directions. The first direction focuses on learning with pre-defined primitive libraries~\cite{chitnis2020efficient,dalal2021accelerating,nasiriany2022augmenting,lee2019learning,9196619,hausknecht2015deep}, where primitives are manually designed and parameterized for policy learning. Another line of research investigates primitive extraction from demonstrations, using either statistical models ~\cite{pastor2009learning,manschitz2015learning,kroemer2015towards,chitnis2020efficient,fox2019hierarchical} or deep neural networks~\cite{gams2018deep,seker2019conditional,gao2024prime,liu2024infocon,zhou2024maxmi,qu2025spatialvla}. Recent works explore hierarchical frameworks~\cite{yu2018one,le2018hierarchical,kipf2019compile,xie2020deep} and model-based methods~\cite{chen2023predicting} integrates primitives for solving stowing tasks, while~\cite{shi2023waypoint} extracts waypoints for motion interpolation. In contrast, our PPL present a novel two-stage framework that leverages primitive prompts to represent shared primitives and finally enables lifelong robot learning.

\section{Problem Formulation}
\label{sec:formulation}
\noindent \textbf{Multi-Skill Pre-Training Stage}:
In multi-skill pre-training stage, 
we formulate the problem as follows. 
Given a set of robot tasks
$C = \{T_j\}_{j=1}^J$, for each task $j$, we have $N$ expert demonstrations $\{\tau_{j,i}\}_{i=1}^N$, where each demonstration $\tau_{j,i}$ is a sequence of state-action pairs. We formulate robot imitation learning as an action sequence prediction problem, aiming to minimize the error in future actions conditioned on historical states. The standard behavioral cloning loss is used to optimize policy $\pi$ over these demonstrations:
\begin{equation}
\hat{\theta} = \min_{\theta} \sum_{k=1}^{K} \mathbb{E}_{s_t,a_t \sim \mathcal{D}_k} \left[ \sum_{t=0}^{l_k} \mathcal{L} \left( \pi(a|s_t, T_k; \theta), a_k^t \right) \right].
\label{eq:optimization}
\end{equation}
where $L$ is a supervised action prediction loss,
$l_k$ is the length of demonstrations for task $T_k$, and $\theta$ refers to the learnable parameters of the network.

\noindent \textbf{Lifelong Learning Stage}:
In this stage, we build upon the shared knowledge acquired during multi-skill pre-training. 
Our objective is to incrementally learn new skills while retaining performance on old ones. The pre-trained policy continues to encounter a sequence of tasks, denoted as ${T_1, ... , T_K}$. For each task $T_k$, the policy receives $N$ demonstrations $D_k = {\tau_k^1, ... , \tau_k^N}$. Here, $\mathcal{D}_k$ only contains data from the current task, and $s_t$ should be interpreted as $s_{\leq t}$.

\section{Methodology}
\label{sec:method}
\begin{figure*}[t]
\vspace{-15pt}
\centering
\includegraphics[width=0.95\textwidth, keepaspectratio, interpolate=true]{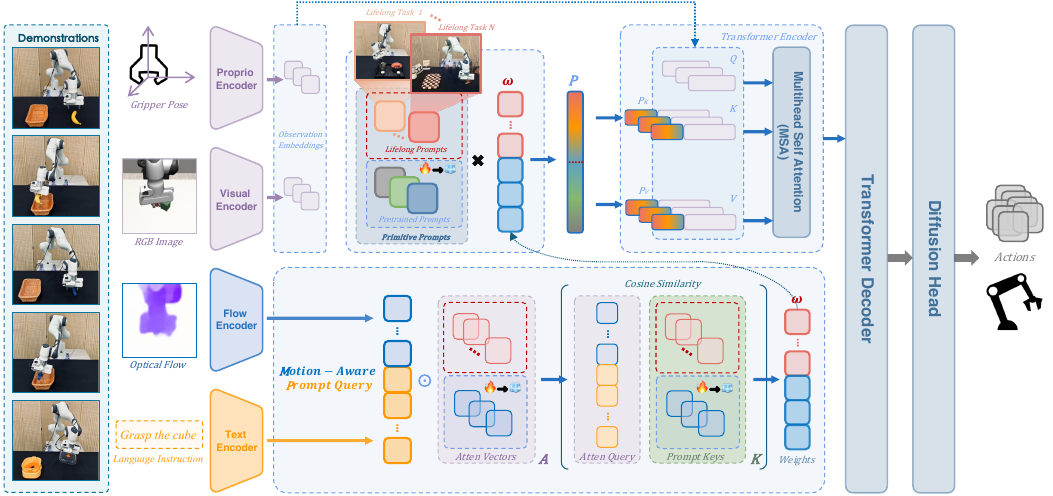}
 \vspace{-5pt}
\caption{\textbf{The overview of Primitive Prompt Learning (PPL)}. In pre-training stage, our input consist of proprioception, image observation and language instruction. A set of primitive prompts are queried via motion-aware query module to obtain a weighted-sum prompt $P$, which is prepended to each layer of diffusion transformer policy. For new skill acquisition with expert demonstrations, new lifelong prompts are concatenated and optimized with frozen pretrained prompts, following the same input/output flow as the pre-training. The notation $\text{fire} \rightarrow \text{ice}$ indicates that prompts are optimized during the pre-training stage and then frozen in the lifelong learning stage.}
\vspace{-10pt}
\label{fig:overview}
\end{figure*}

Fig.~\ref{fig:overview} shows the overview of the proposed PPL.
Given an input demonstration stream $\{D_i\}_{i=1}^J$ and instruction T, we aim to learn a set of reusable and extensible primitive prompts. In Sec.~\ref{sec:prompt}, we present the multi-skill pre-training stage of PPL, where we learn primitive prompts to represent shared primitives across different skills. To effectively capture shared knowledge, we introduce motion-aware prompting in Sec.~\ref{sec:map}, which combines optical flow with task-conditional semantic information to model both semantic and motion-shared primitives. Finally, Sec.~\ref{sec:lifelong} describes the lifelong learning stage of PPL, where new lifelong prompts are concatenated and optimized with frozen pretrained prompts, enabling knowledge transfer from old skills to new ones without requiring access to previous data. 

\vspace{-10pt}

\subsection{Multi-Skill Pre-Training}
\label{sec:prompt}
In the multi-skill pre-training stage of PPL, our goal is to learn a set of reusable and extensible primitive prompts that can effectively capture shared knowledge across multiple robotic skills, enhancing performance across diverse tasks. 
To do this, we apply primitive prompts to the diffusion transformer policy, prepending prompts to the keys and values of the multi-head self-attention (MSA) layers, with distinct prompting parameters for each layer. We define our prompt parameter as $p \in \mathbb{R}^{L_p \times D}$, where $L_p$ represents the prompt length and $D$ denotes the embedding dimension. In a typical MSA layer with input $h \in \mathbb{R}^{L \times D}$, the query, key, and value are represented as $h_Q$, $h_K$, and $h_V$ respectively. The layer's output is computed as follows:
\begin{equation}
\begin{aligned}
&\text{MSA}(h_Q, h_K, h_V) = \text{Concat}(h_1, \ldots, h_m)W^O \\
&\text{where } h_i = \text{Attention}\left(h_QW_i^Q, h_KW_i^K, h_VW_i^V\right)
\end{aligned}
\end{equation}
where $W^O$, $W^Q_i$, $W^K_i$, and $W^V_i$ are projection matrices, and $m$ denotes the number of attention heads.
PPL involves splitting the prompt $p$ into $\{p^K, p^V\} \in \mathbb{R}^{(L_p/2) \times D}$ and prepending these to $h^K$ and $h^V$ using the prefix-prompt method followed by~\cite{li2021prefixtuningoptimizingcontinuousprompts}:
\begin{equation}
f_{P-T}(\mathbf{p}, \mathbf{h}) = \text{MSA}(h_Q, [\mathbf{p}_K; h_K], [\mathbf{p}_V; h_V]).
\end{equation}
Typically, prompt learning methods often rely on task identifiers or language instructions to guide the learning process, involving prepending learnable tokens to the input or intermediate representations of a model. 
However, these approaches that heavily rely on high-level representations face a key limitation: they often struggle to facilitate mutual improvement between tasks that are not semantically similar. This can potentially overlook the rich temporal and motion information inherent in robotic actions. 
For example, while these methods are effective for knowledge transfer between semantically similar tasks like "grasp cube" and "grasp mug", they fall short in capturing shared primitives across semantically distinct but motion-related tasks. This limitation can result in sub-optimal knowledge transfer between seemingly unrelated tasks such as "\textit{grasp mug}" and "\textit{place banana}", which, despite their semantic differences, may share common underlying primitives.


\subsection{Motion-Aware Prompting}
\label{sec:map}

To address these limitations and capture semantic and motion shared primitives across different skills, we propose Motion-Aware Prompting (MAP). MAP combines optical flow with task-conditional semantic information, allowing us to capture and leverage common primitives across seemingly disparate tasks.
Motion-aware optical flow provides a rich representation of motion dynamics within the scene, capturing the essential kinematic properties of primitive actions. This motion-centric approach allows us to identify and learn common movement patterns across skills, even when the high-level semantics differ. 
Additionally, optical flow offers a degree of invariance to appearance changes, focusing instead on the underlying motion structure. This property is particularly valuable in robotics, where the same primitive-level manipulation might be performed on objects with vastly different visual and semantic characteristics.
The optical flow can be represented mathematically as:
\begin{equation}
I(x, y, t) = I(x + u\Delta t, y + v\Delta t, t + \Delta t),
\end{equation}
where $I$ is the image intensity, $(u, v)$ is the optical flow vector, and $\Delta t$ is the time step. 
This formulation is based on the brightness constancy assumption, which posits that the intensity of a moving pixel remains constant over short time intervals. By focusing on pixel movement rather than intensity changes, and by representing motion as a vector $(u, v)$, this captures local motion patterns independent of specific textures or colors.
To estimate optical flow, we employ RAFT~\cite{teed2020raft}, which computes the flow iteratively:
\begin{equation}
f_{k+1} = f_k + \Delta f_k,
\end{equation}
\begin{equation}
\Delta f_k, h_{k+1} = \text{GRU}(C(f_k), h_k),
\end{equation}
where $f_k$ is the flow estimate at iteration $k$, $\Delta f_k$ is the flow update, $C$ is a correlation volume, $h_k$ is a hidden state, and GRU is a gated recurrent unit.

While optical flow provides rich motion information, relying solely on it may cause policy to overlook distinctions between different skills or tasks.  To address this limitation and create a more comprehensive representation, we integrate optical flow with task-specific instructions.  We embed these instructions by pre-trained CLIP~\cite{radford2021learning} model, complementing the motion information from optical flow.  This integration allows MAP to leverage both semantic understanding and low-level motion features, enabling the model to identify shared primitives across different skills while maintaining task-specific knowledge, thus promoting effective knowledge transfer between tasks.
The formalized representation of motion-aware prompt query is:
\begin{equation}
\text{MAP}(T, F) = f_{\text{prompt}}(E_{\text{CLIP}}(T), \Phi(F)),
\end{equation}
where $T$ is the task description, $F$ is the optical flow from RAFT, $E_{\text{CLIP}}(T)$ is the CLIP-based semantic embedding function, $\Phi(F)$ is the flow feature extraction function, and $f_{\text{prompt}}$ is a learned function that combines semantic and motion information.
The CLIP-based semantic embedding ensures task-specificity,
while the flow feature enables fine-grained primitive modeling. 
MAP enables policy to model and transfer primitives, thereby facilitating mutual improvement and lifelong expansion across diverse skills.

\subsection{Lifelong Skill Acquisition}
\label{sec:lifelong}
In the lifelong learning stage of PPL, we introduce a lifelong skill acquisition method by concatenating new primitive prompts with frozen pretrained prompts, enabling knowledge transfer from old skills to new ones. Specifically, we introduce a set of prompt components $P \in \mathbb{R}^{M \times D}$, where M is the number of components. Each prompt component P initializes attention vectors $A \in \mathbb{R}^{D \times M}$, keys $K \in \mathbb{R}^{M \times D}$. PPL combines these components through the following steps:
\begin{equation}
\text{Atten\_Query} = \text{MAP}(T, F) \odot \mathbf{A},
\end{equation}
where $\text{MAP}(T, F)$ is a text-flow query function and $\odot$ denotes the Hadamard product. Then, to determine the appropriate weighting vector for each skill, we compute the weight vector $\alpha$ based on the similarity between a text-flow query and a set of keys associated with the prompt components:
\begin{equation}
\alpha_m = cos\_sim(\text{Atten\_Query}, \mathbf{K}_m),
\end{equation}
where $cos\_sim$ represents a similarity function, and $\mathbf{K}_m$ is the key for the $m$-th prompt component.
\begin{equation}
p = \sum_{m} \alpha_m P_m.
\end{equation}
Here, we employ a weighted summation of prompt components, $P_m$ is the $m$-th prompt component. 

As shown in Fig.~\ref{fig:overview}, to achieve lifelong skill acquisition, we freeze the primitive prompts obtained after the stage of multi-skill pre-training, ensuring that the expanded parameters do not alter the weight $\alpha$ calculations for previously learned tasks during the lifelong learning stage.
When acquiring new skills, we expand P, K, and A, increasing their dimensions to $P \in \mathbb{R}^{Z \times D}$, $A \in \mathbb{R}^{D \times Z}$, and $K \in \mathbb{R}^{Z \times D}$, where $Z$ represents the number of prompt components after incorporating new skills, and $D$ is the embedding dimension.
Notably, while training the $t$-th lifelong task, we only update the parameters of the lifelong prompts for this specific task. Nevertheless, when calculating weights, we still involve all frozen primitive prompts and lifelong prompt components from all previous $t$ tasks.
\begin{algorithm}
\scriptsize
\caption{PPL: Primitive Prompt Learning}
\begin{algorithmic}[1]
\Require Visual demonstrations $\{D_i\}_{i=1}^J$, Skill descriptions $T$
\Ensure Learned Primitive Prompts

\State Initialize $p \in \mathbb{R}^{L_p \times D}$ \Comment{Initialize primitive prompts}
\For{each skill $j$ in $\{1, \ldots, J\}$}
    \State $f_{k+1} = f_k + \Delta f_k$ \Comment{Compute optical flow using RAFT}
    \State $\text{MAP}(T, F) = f_\text{prompt}(E_\text{CLIP}(T), \Phi(F))$ \Comment{Motion-Aware Prompting}
    \State $f_{P-T}(p, h) = \text{MSA}(h_Q, [p_K; h_K], [p_V; h_V])$ \Comment{Apply prefix-prompt learning}
    \State Compute diffusion loss $\mathcal{L}$ \Comment{Using diffusion transformer policy}
    \State Update $p$ and model parameters to minimize $\mathcal{L}$
\EndFor

\For{each new skill $k$}
    \State Initialize $P \in \mathbb{R}^{M' \times D }$ \Comment{Initialize new prompt components}
    \State Compute $\text{MAP(T, F)}_k$ \Comment{Compute MAP query for new skill}
    \State $\alpha = cos\_sim(q(x) \odot A, K)$ \Comment{Compute attention-based weighting}
    \State $p = \sum_m' \alpha_m' P_m'$ \Comment{Generate new prompt}
    \State Compute diffusion loss $\mathcal{L}$ for new skill \Comment{Using diffusion transformer policy}
    \State Update $p$ and model parameters to minimize $\mathcal{L}$ 
    \State Add $p$ to existing prompts \Comment{Expand prompt set}
\EndFor

\State \textbf{return} Learned prompts
\end{algorithmic}
\end{algorithm}

\vspace{-10pt}

\begin{figure*}[t]
\vspace{-10pt}
\centering
\includegraphics[width=\textwidth, keepaspectratio, interpolate=true]{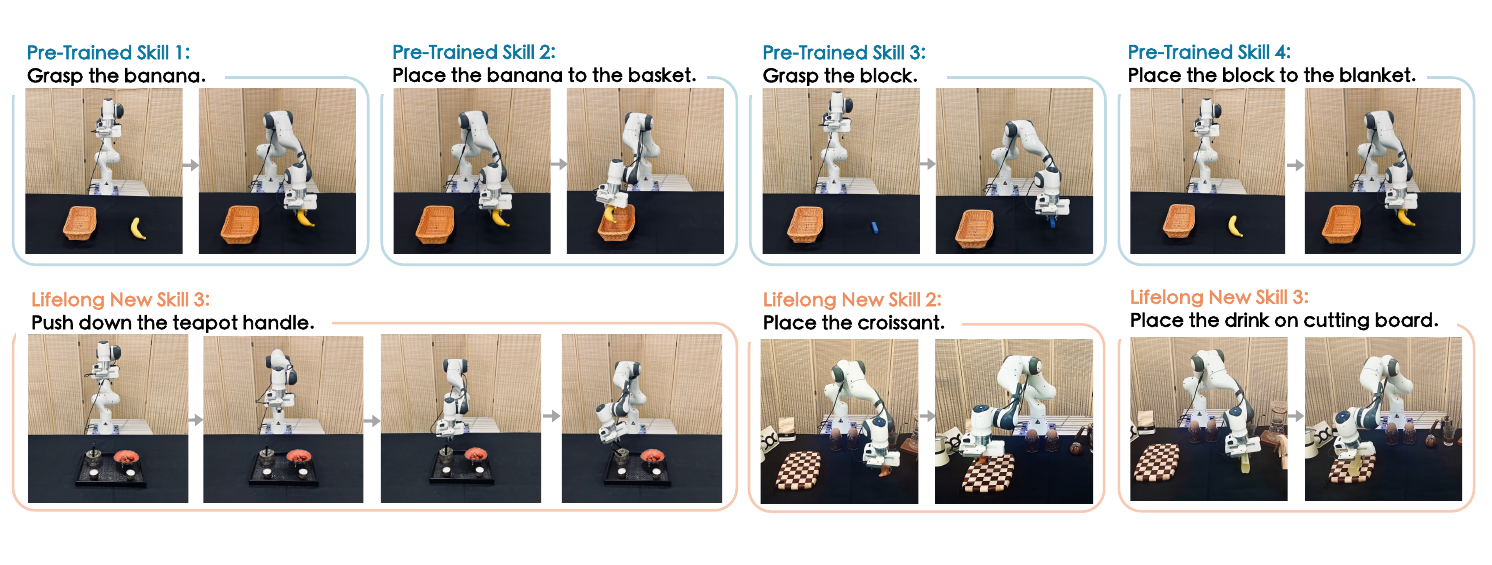}
\vspace{-30pt}
\caption{\textbf{Real-world robot setting.} We proposed 9 real-world skills, 4 of which are used in the pre-training stage and 5 in the lifelong stage, covering a variety of action spaces such as grasp, place, push, and a variety of different objects and distributions.}
\label{fig:franka}
\vspace{-10pt}
\end{figure*}

\begin{figure*}[t]
\centering
\includegraphics[width=0.95\textwidth, keepaspectratio, interpolate=true]{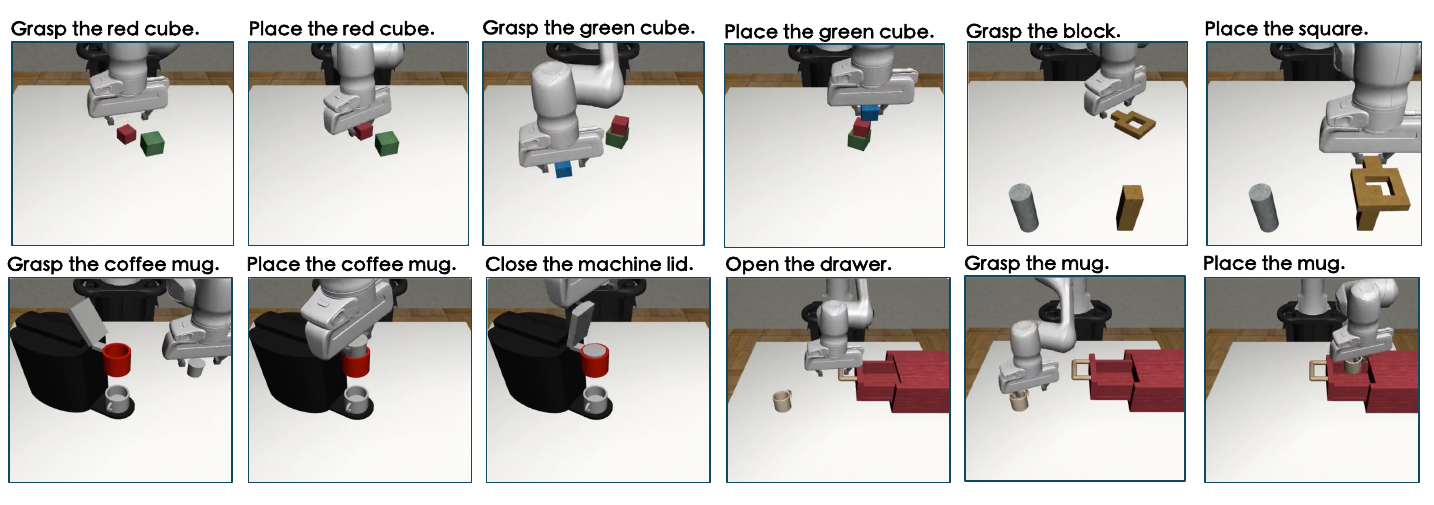}
\vspace{-10pt}
\caption{\textbf{Illustration of our skill dataset.} Our skill dataset is constructed based on MimicGen and LEBRO benchmark with diverse action spaces and scene variations.}
\label{fig:mimicgen}
\vspace{-10pt}
\end{figure*}


\section{Experiments}
\label{sec:result}
\subsection{Experimental Setup}

\noindent\textbf{Simulation tasks.} We conduct our simulation experiments by a large-scale skill dataset that constructed based on MimicGen~\cite{mandlekar2023mimicgen} and LIBERO~\cite{liu2023libero}. In our skill dataset, each skill is associated with its language instruction. For example, a skill might be “Grasp the mug" or "Open the drawer". As shown in Fig.~\ref{fig:mimicgen}, 
our dataset incorporates skills from MimicGen, 
each containing 1K human demonstrations and with broad initial state distributions, effectively showing the generalization for multitask evaluation. 
We also include skills from LIBERO, a lifelong robotic manipulation benchmark.
By building our large-scale skill dataset, we ensure a comprehensive range of robotic manipulation scenarios, enabling PPL to be applied to diverse and challenging tasks. 

\noindent\textbf{Real-world experiments.} The real-robot experiments are conducted on the Franka Panda robotic arm. As shown in Fig.~\ref{fig:franka}, we perform multitask pre-training on four distinct skills, each comprising 200 human demonstrations with broad initial state distributions. To evaluate our policy's ability for lifelong learning, we conduct training and validation on four additional skill tasks. The objects involved in these tasks, such as banana, block, and various utensils, are randomly placed to assess position generalization. All metrics are evaluated with 15 independent runs for each skill, ensuring robust performance assessment across different initial conditions and task variations.


\begin{figure*}[t]
\vspace{-10pt}
    \centering
    \includegraphics[width=\textwidth, keepaspectratio, interpolate=true]{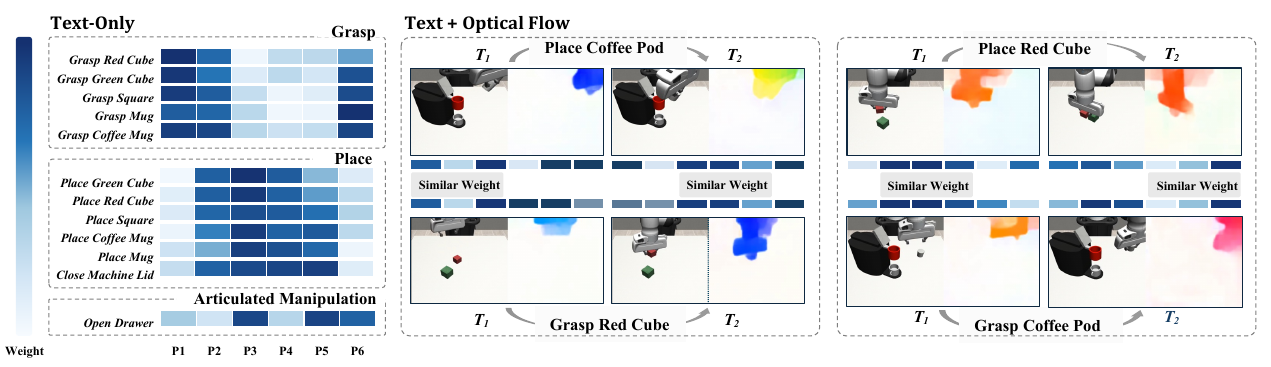}
    \vspace{-20pt}
    \caption{\textbf{Impact of Motion-Aware Prompt Query on Prompt Weights.} This figure illustrates the weight distributions when using only text as the query (left) and when using our text-flow query (right). When only text is used as the prompt query, the weight responses exhibit similarities only in semantically related tasks. In contrast, our text-flow query enables the policy to have similar weight responses even in semantically different skills, allowing different skills to learn primitives in the latent space.}
    \label{fig:weight}
\vspace{-10pt}
\end{figure*}

\noindent\textbf{Evaluation Metrics.} Following~\cite{liu2023tail}, we employ Forward Transfer Weight (FWT) and Backward Transfer Weight (BWT) to evaluate the performance of lifelong learning. FWT is computed by the maximum success rate our policy can achieve when adapting to a new task. We denote FWT at task $k$ as $F_k$. Meanwhile, BWT measures the success rate increase on previous tasks. Specifically, when adapting to the $k$-th task, we first record the best FWT model on this task and then evaluate this model on all previous $k - 1$ tasks, obtaining success rate $S_i$, $1 \leq i \leq k - 1$. Then we compute the success rate difference between the new model and the best FWT of the previous $k - 1$ tasks and then average among them to obtain the BWT metric:
\begin{equation}
B = \frac{1}{k-1}\sum_{i=1}^{k-1} (S_i - F_i),
\end{equation}
For both FWT and BWT metrics, higher values indicate better performance in terms of knowledge transfer and retention across tasks.

\subsection{Multi-Skill Pre-Training}
As shown in Tab.\ref{tab:comprehensive-comparison-pretraining-lifelong}, our PPL achieves the highest success rates across all pre-training tasks in the LIBERO-GOAL environment. Compared to the MoE-based policy~\cite{wang2024sparse}, PPL improves the average success rate across all tasks by 17\%. Appendix provides a detailed success rates for each task within MimicGen benchmark. We further evaluate PPL's ability to learn generalizable cross-skill information in real-world scenarios. Tab. \ref{tab:comprehensive-comparison-pretraining-lifelong} presents the results of real-world experiments, where PPL consistently outperforms existing approaches. These results validate PPL's effectiveness in both simulated and real-world environments.

\subsection{Lifelong Learning}
For lifelong learning tasks, we conducted a comparative analysis of PPL against traditional sequential learning approaches, experience replay-based methods, and task-specific LoRA~\cite{liu2023tail}. As illustrated in Tables \ref{tab:comprehensive-comparison-7tasks} and \ref{tab:comprehensive-comparison-pretraining-lifelong} , PPL demonstrates superior performance in simulated environments, achieving state-of-the-art performance in both FWT and BWT metrics. Furthermore, Tab.\ref{tab:comprehensive-comparison-pretraining-lifelong} presents evidence that in real-world scenarios, PPL not only facilitates the acquisition of cross-skill primitives during the pre-training phase but also effectively leverages this pretrained primitives into the new skill acquisition stage, enabling shared knowledge transfer from old skills to new ones. Notably, PPL surpasses existing approaches without requiring access to replay experiences.

\begin{table*}[htbp]
\centering
\tiny
\resizebox{0.8\textwidth}{!}{ 
\begin{tabular}{l|cc|cc|c|c}
\hline
\multirow{4}{*}{Task} & \multicolumn{4}{c|}{Conventional Methods} & \multicolumn{2}{c}{Adapter-based Methods} \\
\cline{2-7}
 & \multicolumn{2}{c|}{Sequential} & \multicolumn{2}{c|}{ER} & LoRA & PPL (Ours) \\
 & FWT $\uparrow$ & BWT $\uparrow$ & FWT $\uparrow$ & BWT $\uparrow$ & FWT $\uparrow$ & FWT $\uparrow$ \\
\hline
Task 1 & 0.87 $\pm$ 0.07 & - & 0.79 $\pm$ 0.12 & - & \textbf{0.89} $\pm$ 0.02 & 0.88 $\pm$ 0.00 \\
Task 2 & 0.73 $\pm$ 0.07 & -0.57 $\pm$ 0.08 & 0.71 $\pm$ 0.07 & -0.23 $\pm$ 0.08 & \textbf{0.79 $\pm$ 0.01} & 0.75 $\pm$ 0.12 \\
Task 3 & 0.79 $\pm$ 0.04 & -0.48 $\pm$ 0.12 & 0.67 $\pm$ 0.07 & -0.37 $\pm$ 0.11 & 0.81 $\pm$ 0.07 & \textbf{0.83 $\pm$ 0.03} \\
Task 4 & 0.77 $\pm$ 0.03 & -0.62 $\pm$ 0.17 & 0.64 $\pm$ 0.07 & -0.44 $\pm$ 0.19 & 0.78 $\pm$ 0.00 & \textbf{0.79 $\pm$ 0.02} \\
Task 5 & 0.49 $\pm$ 0.07 & -0.69 $\pm$ 0.24 & 0.35 $\pm$ 0.14 & -0.57 $\pm$ 0.23 & \textbf{0.62 $\pm$ 0.12} & 0.60 $\pm$ 0.09 \\
Task 6 & 0.64 $\pm$ 0.12 & -0.66 $\pm$ 0.24 & 0.52 $\pm$ 0.19 & -0.61 $\pm$ 0.23 & 0.61 $\pm$ 0.12 & \textbf{0.73 $\pm$ 0.14} \\
Task 7 & 0.32 $\pm$ 0.05 & -0.69 $\pm$ 0.18 & 0.11 $\pm$ 0.00 & -0.58 $\pm$ 0.24 & 0.43 $\pm$ 0.26 & \textbf{0.54 $\pm$ 0.11} \\
\hline
Average & 0.65 $\pm$ 0.06 & -0.56 $\pm$ 0.16 & 0.61 $\pm$ 0.09 & -0.46 $\pm$ 0.18 & 0.78 $\pm$ 0.09 & \textbf{0.83 $\pm$ 0.03} \\
\hline
\end{tabular}
} 
\caption{\textbf{Lifelong Performances with MimicGen.} PPL achieved the best success rate in both multi-skill pre-training and lifelong learning, as well as demonstrating superior lifelong learning capabilities.}
\label{tab:comprehensive-comparison-7tasks}
\vspace{-10pt}
\end{table*}

\begin{table}
\vspace{5pt}
\centering
\small
\resizebox{\linewidth}{!}{ 
\begin{tabular}{l|ccc}
\hline
\multirow{2}{*}{Task} & \multicolumn{3}{c}{Methods} \\
\cline{2-4}
 & Diffusion-Transformer & MOE & PPL (Ours) \\
\hline
\multicolumn{4}{c}{Multi-Skill Policy Pre-Training} \\
\hline
Pretrain Task 1 & 0.60 $\pm$ 0.05 & 0.82 $\pm$ 0.04 & \textbf{0.99 $\pm$ 0.03} \\
Pretrain Task 2 & 0.25 $\pm$ 0.06 & 0.78 $\pm$ 0.05 & \textbf{0.62 $\pm$ 0.02} \\
\hline
Average & 0.42 $\pm$ 0.09 & 0.73 $\pm$ 0.08 & \textbf{0.84 $\pm$ 0.05} \\
\hline
\multicolumn{4}{c}{Lifelong Learning} \\
\hline
Task & Sequential & ER & PPL (Ours) \\
\hline
Lifelong Task 1 & 0.60 $\pm$ 0.08 & 0.65 $\pm$ 0.07 & \textbf{0.72 $\pm$ 0.04} \\
Lifelong Task 2 & 0.55 $\pm$ 0.09 & 0.58 $\pm$ 0.08 & \textbf{0.68 $\pm$ 0.05} \\
Lifelong Task 3 & 0.50 $\pm$ 0.10 & 0.52 $\pm$ 0.09 & \textbf{0.63 $\pm$ 0.06} \\
\hline
Average & 0.55 $\pm$ 0.09 & 0.58 $\pm$ 0.08 & \textbf{0.68 $\pm$ 0.05} \\
\hline
\end{tabular}
}
\caption{\textbf{Performances with real-world robot tasks.} PPL achieved the best success rate in both multi-skill pre-training stage, as well as demonstrating superior lifelong learning capabilities.}
\label{tab:comprehensive-comparison-pretraining-lifelong}
\vspace{-5pt}
\end{table}

\begin{figure}[!ht]
\centering
\tiny
\resizebox{\linewidth}{!}{
\begin{tabular}{l|ccc}
\hline
\multirow{2}{*}{Task} & \multicolumn{3}{c}{Methods} \\
\cline{2-4}
 & Diff-T & MOE & PPL (Ours) \\
\hline
\multicolumn{4}{c}{Multi-Skill Pre-Training} \\
\hline
Pretrain Task 1 & 0.79 $\pm$ 0.05 & 0.83 $\pm$ 0.04 & \textbf{0.85 $\pm$ 0.03} \\
Pretrain Task 2 & 0.83 $\pm$ 0.11 & 0.85 $\pm$ 0.03 & \textbf{0.86 $\pm$ 0.02} \\
Pretrain Task 3 & 0.84 $\pm$ 0.07 & 0.86 $\pm$ 0.08 & \textbf{0.86 $\pm$ 0.01} \\
Pretrain Task 4 & 0.63 $\pm$ 0.08 & 0.74 $\pm$ 0.07 & \textbf{0.80 $\pm$ 0.03} \\
\hline
Average & 0.77 $\pm$ 0.02 & 0.82 $\pm$ 0.03 & \textbf{0.84 $\pm$ 0.02} \\
\hline
\multicolumn{4}{c}{Lifelong Learning} \\
\hline
Task & Sequential & ER & PPL (Ours) \\
\hline
Lifelong Task 1 & 0.77 $\pm$ 0.08 & 0.73 $\pm$ 0.04 & \textbf{0.78 $\pm$ 0.04} \\
Lifelong Task 2 & 0.65 $\pm$ 0.03 & 0.61 $\pm$ 0.12 & \textbf{0.68 $\pm$ 0.09} \\
Lifelong Task 3 & \textbf{0.74 $\pm$ 0.11} & 0.62 $\pm$ 0.08 & 0.71 $\pm$ 0.06 \\
\hline
Average & 0.72 $\pm$ 0.04 & 0.65 $\pm$ 0.03 & \textbf{0.73 $\pm$ 0.03} \\
\hline
\end{tabular}
\vspace{-10pt}
}
\caption{\textbf{Performances with LIBERO.} When dealing with different tasks in the same scene, PPL achieves the best performance.}
\label{tab:comprehensive-comparison-pretraining-lifelong}
\vspace{-10pt}
\end{figure}


\subsection{Ablation Studies}
\paragraph{Effect of Motion-Aware Prompting}
To validate the effectiveness of our motion-aware prompt query, we visualize the weight distributions when using only text embeddings as query and when using our text-flow query. As shown in the Fig.~\ref{fig:weight}, if only task-specific instruction embedding is used as the prompt query, the weight responses will only exhibit similarities in semantically related tasks, and within a single task, the weights remain the same at each time step. 
In contrast, our text-flow query can capture both semantic and motion-shared primitives across different skills, even those skills are semantically distinct, ultimately enabling primitive representation through primitive prompts.

\begin{figure}[!h]
\centering
\includegraphics[width=1.0\linewidth, keepaspectratio, interpolate=true]{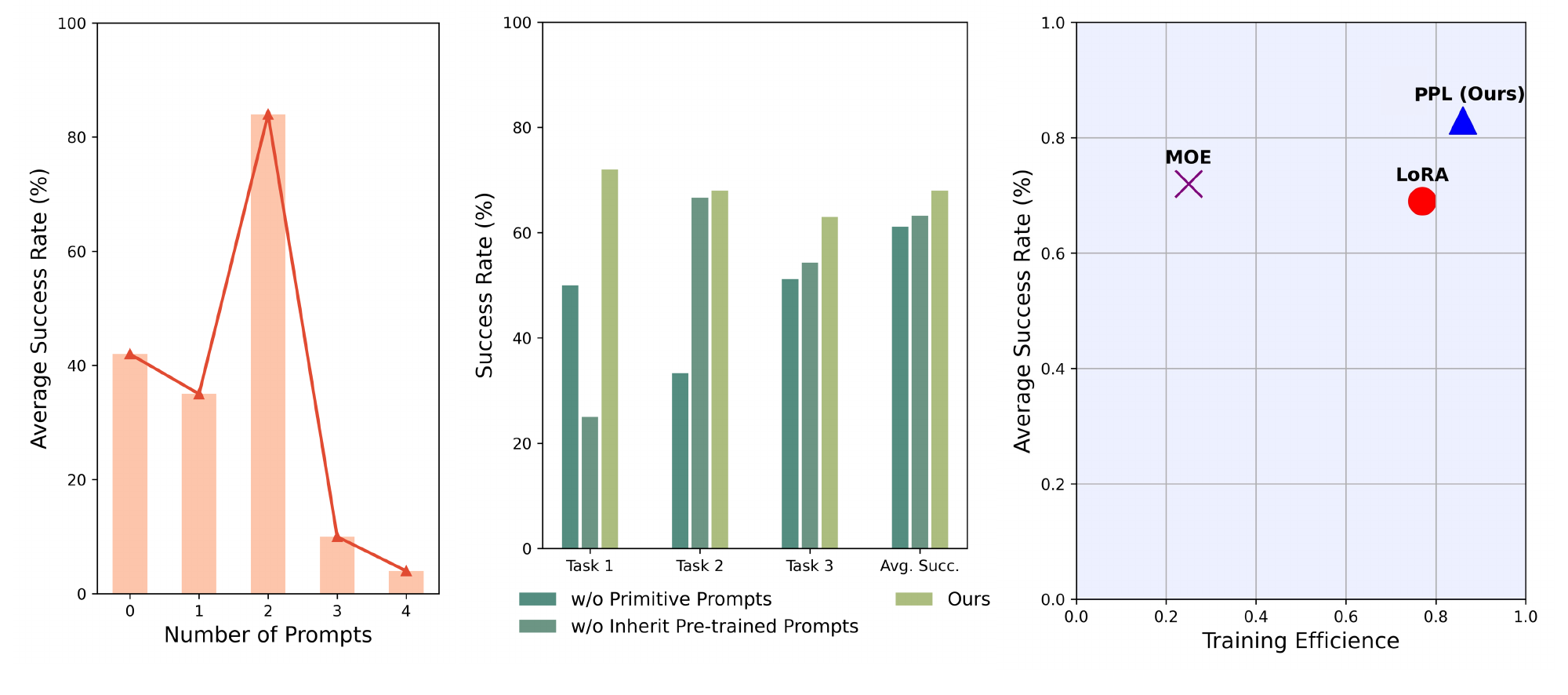}
\vspace{-20pt}
\caption{\textbf{Illustration of ablation studies.} We conducted ablation analysis on the effect of proposed prompt mechanism, the number of prompts, and comparisons with MoE and LoRA.}
\vspace{-15pt}
\label{fig:none}
\end{figure}

\noindent\textbf{Effect of Primitive Prompt Count.}
We conducted a comprehensive investigation into the optimal selection of prompt count during the multi-skill learning.  As various skills undergo joint optimization, primitives are modeled by primitive prompts.  For any specific task, only a subset of primitive prompts responds and matches to extract relevant prior knowledge, while unmatched primitive prompts may introduce noise.  Consequently, as illustrated in Fig. \ref{fig:none}, an increase in the number of prompts does not necessarily correlate with improved performance.  Simultaneously, an insufficient number of prompts may fail to encompass all primitives, underscoring the importance of achieving an appropriate balance in prompt count.

\noindent\textbf{Effect of Primitive Prompt.}
As illustrated in Fig.\ref{fig:none}, significant performance degradation is observed when learning new skills under two conditions: (1) when prompt learning of primitives is omitted during the pre-training phase, or (2) when pre-trained prompts are not utilized in the acquisition of new skills. These findings substantiate the effectiveness of our proposed prompt mechanism in extracting common knowledge from pre-trained skills. Moreover, they demonstrate the mechanism's capacity to repurpose this knowledge during the lifelong learning phase, thereby enhancing the performance of newly acquired skills.

\begin{figure}[!t]
    \vspace*{-2mm}
    \centering
\includegraphics[width=1.0\linewidth, keepaspectratio,]{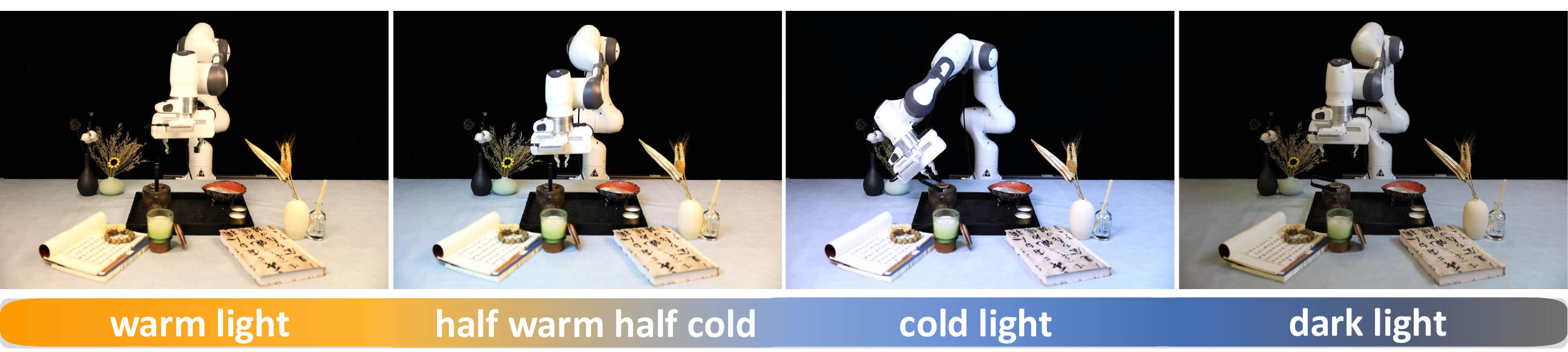}
    \vspace*{-7mm}
    \caption{\footnotesize Visualization of Franka robot arm pushing the tea pot handle aside under light variation level L3 (warm→cold→dark).}
    \label{fig:light}
\end{figure}

\noindent\textbf{Robustness Analysis under Complex lighting Conditions.}
As shown in Fig.\ref{fig:light}, we design a challenging scenario under varying lighting conditions. The experiment simulates a dynamic lighting environment where the illumination transitions from warm to cold, and finally to dark during a single skill execution. 
This setup creates a more demanding visual challenge for the robot, as it must adapt to rapidly changing lighting conditions while performing the task of pushing the tea pot handle aside. Such extreme variations in lighting are rarely encountered in typical real-world scenarios, making this a particularly stringent test of PPL's robustness.
According to Tab.\ref{tab:light_tab}, while both Diffusion Policy and our PPL with flow-based MAP degrade under these conditions, using only text as PPL's prompt query shows improved performance. To address this limitation, our future work will explore incorporating depth information and 3D scene flow for further robustness.


\subsection{Discussion on PPL v.s. Lora and MoE}
\vspace{-5pt}
Recently, some studies have explored the effectiveness of LoRA~\cite{liu2023tail} and MOE~\cite{wang2024sparse} in enhancing lifelong robot learning. However, as illustrated in Fig. \ref{fig:none}, our experiments demonstrate that although MOE excels in terms of average success rate, its training speed is slower due to the additional computational overhead introduced by its gating network and multiple expert networks. MOE's training time is approximately twice that of LoRA and our proposed method. LoRA, on the other hand, emerges as the frontrunner in terms of training speed, while its overall performance falls short of its competitors. Notably, PPL achieves performance surpassing that of MOE while maintaining comparable training speed. This balance of efficiency and efficacy enables PPL to effectively combine the strengths of LoRA and MOE, facilitating faster skill knowledge acquisition while preserving high performance.

\begin{table}[!t]
\centering
\caption{\footnotesize \textbf{\footnotesize Quantitative Results under Various Lighting Conditions.} \\
\quad 
\scriptsize * Light Variation Level: warm (L1), warm→cold (L2), warm→cold→dark (L3); \quad † Static Light Condition: Constant light with different colors.}
\vspace*{-3mm}
{\scriptsize
\renewcommand{\arraystretch}{0.8}
\newcolumntype{C}{>{\centering\arraybackslash}X}
\begin{tabularx}{\linewidth}{@{}l*{6}{C}@{}}
\toprule
\multirow{2}{*}{Methods} & \multicolumn{3}{c}{Light Variation Level*} & \multicolumn{3}{c}{Static Light Condition†} \\
\cline{2-7}
 & \raisebox{-0.5ex}{L1} & \raisebox{-0.5ex}{L2} & \raisebox{-0.5ex}{L3} & \raisebox{-0.5ex}{warm} & \raisebox{-0.5ex}{cold} & \raisebox{-0.5ex}{dark} \\
\midrule
Diff-T & 0.82 & 0.80 & 0.75 & 0.82 & 0.85 & 0.52 \\
PPL w flow+text & \textbf{0.83} & 0.76 & 0.61 & \textbf{0.83} & \textbf{0.87} & 0.48\\
PPL w text & 0.81 & \textbf{0.84} & \textbf{0.80} & 0.81 & 0.86 & \textbf{0.61} \\
\bottomrule
\end{tabularx}}
\label{tab:light_tab}
\end{table}


\vspace{-5pt}
\section{Conclusion}
\label{sec:conclusion}
\vspace{-5pt}
In this work, we present Primitive Prompt Learning for lifelong robotic skill learning. Motion-aware prompts via text flow query mechanism are proposed to learn reusable and extensible primitive prompts across multiple skills and achieve superior results in multi-skill pre-training. Moreover, for new skill acquisition, lifelong prompts 
are concatenated and optimized with frozen pretrained prompt, enabling
knowledge transfer between old and new skills.
Finally, we construct a large-scale skill dataset and demonstrate the superior perform of PPL in multi-skill pre-training and lifelong skill acquisition.

\vspace{10pt}

\noindent\textbf{Acknowledgements.} This work is supported by the Shanghai AI Laboratory, the National Natural Science Foundation of China (62376222), Young Elite Scientists Sponsorship Program by CAST (2023QNRC001).

{
    \small
    \bibliographystyle{ieeenat_fullname}
    \bibliography{main}
}


\end{document}